\documentclass[a4paper,10pt]{article}
\usepackage[utf8]{inputenc}
\usepackage{hyperref}
\usepackage{amsfonts}

\title{A Note on Kaldi's PLDA Implementation}
\author{Ke Ding \\ kedingpku@gmail.com}

\begin{document}

\maketitle

\href{https://github.com/kaldi-asr/kaldi}{Kaldi}\cite{kaldi}’ s PLDA  implementation is based on \cite{plda}, the so-called two-covariance PLDA by \cite{pldareview}. The authors derive a clean update formula for the EM training and give a detailed comment in the \href{https://github.com/kaldi-asr/kaldi/blob/master/src/ivector/plda.cc#L427}{source code}. Here we add some explanations to make formula derivation easier to catch.

\section{Background}

Recall that PLDA assume a two stage generative process:

1) generate the class center according to 
\begin{equation}
y \sim \mathcal{N}(\mu, \Phi_b)
\end{equation}
2) then, generate the observed data by:
\begin{equation}
x \sim \mathcal{N}(y, \Phi_w)
\end{equation}
where $\mu$ is estimated by the global mean value:
\begin{equation}
\mu = \frac{1}{N} \sum_{k=1}^K \sum_{i=1}^{n_k} z_{ki}
\end{equation}
where $z_{ki}$ depicts the $i$-th sample of the $k$-th class.

So let's turn to the estimation of $\Phi_b$ and $\Phi_w$. 

Note that, as $\mu$ is fixed, we remove it from all samples. Hereafter, we assume all samples have pre-processed by removing $\mu$ from them.

The prior distribution of an arbitrary sample $z$  is:
\begin{equation}
p(z) \sim \mathcal{N}(0,  \Phi_w + \Phi_w)
\end{equation}
Let's suppose the mean of a particular class is $m$, and suppose that that class had $n$ examples. Then
\begin{equation}
m = \frac{1}{n}\sum_{i=1}^n z_i \sim \mathcal{N}(0, \Phi_w + \frac{\Phi_w}{n} )    
\end{equation}
i.e. $m$ is Gaussian-distributed with zero mean and variance equal to the between-class variance plus $1/n$ times the within-class variance. Now, $m$ is observed (average of all observed samples).  

\section{EM}
We're doing an E-M procedure where we treat $m$ as the sum of two variables:
\begin{equation}
m = x + y
\end{equation}
where $x \sim N(0, \Phi_b)$, $y \sim N(0, \Phi_w/n)$.
The distribution of $x$ will contribute to the stats of $\Phi_b$, and $y$ to $\Phi_w$.

\subsection{E Step}
Note that given $m$, there's only one latent variable in effect. Observe that $y = m - x$, so we can focus on working out the distribution of $x$ and then we can very simply get the distribution of $y$.

Given $m$, the posterior distribution of $x$ is:
\begin{equation}
p(x|m) = \int_y p_x(x|m, y)p_y(y)\mathrm{d}y = p_x(x|m)p_y(m-x|m)
\end{equation}
Hereafter, we drop the condition on $m$ for brevity.
\begin{equation}
p(x) = p_x(x)p_y(m - x) =\mathcal{N}(x|0, \Phi_b) \mathcal{N}(x|m, \Phi_w/n)
\end{equation}

Since two Gaussian's product is Gaussian as well, we get. 
\begin{equation}
p(x) = \mathcal{N}(w, \hat \Phi)
\end{equation}
where $\hat \Phi = (\Phi_b^{-1} + n \Phi_w^{-1}) ^{-1}$ and  $w = \hat \Phi  n\Phi_w^{-1} m$.

$\hat \Phi$ and $w$ can be inferred by comparing the first and second order coefficients to the standard form of log Gaussian. As Kaldi's comment does\footnote{The C is different from line to line}:
\begin{eqnarray}
\ln p(x) = C - 0.5 (x^T \Phi_b^{-1} x + (m-x)^T n\Phi_w^{-1}(m-x)) = \\
\nonumber C - 0.5 x^T (\Phi_b^{-1} + n\Phi_w^{-1}) x + x^T z
\end{eqnarray}
where $z = n \Phi_w^{-1} m$, and we can write this as:
\begin{equation}
\ln p(x) = C - 0.5 (x-w)^T (\Phi_b^{-1} + n \Phi_w^{-1}) (x-w)
\end{equation}
where $x^T (\Phi_b^{-1} + n \Phi_w^{-1}) w = x^T z$, i.e. 
\begin{equation}
(\Phi_b^{-1} + n \Phi_w^{-1}) w = z = n \Phi_w^{-1} m
\end{equation}
so
\begin{equation}
w = (\Phi_b^{-1} + n \Phi_w^{-1})^{-1} * n \Phi_w^{-1} m
\end{equation}
\begin{equation}
\hat \Phi = (\Phi_b^{-1} + n \Phi_w^{-1}) ^{-1}
\end{equation}

\subsection{M Step}
The objective function of EM update is:
\begin{eqnarray}
Q = \mathbb{E}_x \ln p_x(x) = \mathbb{E}_x -0.5 \ln|\Phi_b| -0.5 x^T (\Phi_b)^{-1} x\\  
\nonumber = -0.5 \ln|\Phi_b| -0.5 \mathrm{tr}( xx^T (\Phi_wb)^{-1})
\end{eqnarray}
derivative w.r.t $\Phi_b$ is as follows:
\begin{equation}
\frac{\partial }{\partial (\Phi_b)} = -0.5 (\Phi_b)^{-1} + 0.5  (\Phi_b)^{-1} \mathbb{E}[xx^T]  (\Phi_b)^{-1}
\end{equation}
To zero it, we have:

\begin{equation}
\hat \Phi_b = \mathbb{E}_x[xx^T] = \hat \Phi + \mathbb{E}_x[x]  \mathbb{E}_x[x] ^T = \hat \Phi +ww^T
\end{equation}
Similarly, we have:
\begin{equation}
\hat \Phi_w/n = \mathbb{E}_y[yy^T] = \hat \Phi + \mathbb{E}_y[y]  \mathbb{E}_y[y] ^T = \hat \Phi +(w-m)(w-m)^T
\end{equation}

\section{Summary}
Recap that given samples of certain class, we can calculate the following statistics:
\begin{equation}
\hat \Phi = (\Phi_b^{-1} + n \Phi_w^{-1}) ^{-1}
\end{equation}
\begin{equation}
w =\hat  \Phi  n \Phi_w^{-1} m
\end{equation}
\begin{equation}
\hat \Phi_w = n( \Phi +(w-m)(w-m)^T)
\end{equation}
\begin{equation}
\hat \Phi_b =  \hat \Phi + ww^T
\end{equation}
Given $K$ classes, updated estimation via EM will be:
\begin{equation}
\Phi_w = \frac{1}{K}\sum_k n_k(\hat  \Phi_k +(w_k-m_k)(w_k-m_k)^T)
\end{equation}
\begin{equation}
\Phi_b = \frac{1}{K}\sum_k (\hat \Phi_k + w_kw_k^T)
\end{equation}

Finally, Kaldi use the following update formula for $\Phi_w$\footnote{$S$ is not the result of EM used here, since $m = x + y$ only take pooling of data into consideration.}: 
\begin{equation}
\Phi_w = \frac{1}{N} (S + \sum_k n_k(\hat  \Phi_k +(w_k-m_k)(w_k-m_k)^T))
\end{equation}

where $S$ is the scatter matrix $S = \sum_k \sum_i (z_{ki} - c_k)$, and $c_k = \frac{1}{n_k}\sum_i z_{ki}$ is the mean  of samples of the $k$-th class.

For other EM training methods for PLDAs, see \cite{pldareview} and the references herein.



\begin{thebibliography}{9}
\bibitem{kaldi} Dan et al. \textit{The Kaldi Speech Recognition Toolkit}. ASRU. 2011.
\bibitem{plda} Ioffe. \textit{Probabilistic Linear Discriminant Analysis}. ECCV. 2006.
\bibitem{pldareview} Sizov et al. \textit{Unifying Probabilistic Linear Discriminant Analysis Variants in Biometric Authentication}. SSSPR. 2014. 
\end{thebibliography}
\end{document}